# Analyzing and Mitigating the Impact of Permanent Faults on a Systolic Array Based Neural Network Accelerator

Jeff (Jun) Zhang    Tianyu Gu    Kanad Basu    Siddharth Garg
*Department of Electrical and Computer Engineering, New York University, Brooklyn, New York, 11201*
Email: {jeffjunzhang,tg1553,kb150, sg175}@nyu.edu

*Abstract*—Due to their growing popularity and computational cost, deep neural networks (DNNs) are being targeted for hardware acceleration. A popular architecture for DNN acceleration, adopted by the Google Tensor Processing Unit (TPU), utilizes a systolic array based matrix multiplication unit at its core. This paper deals with the design of fault-tolerant, systolic array based DNN accelerators for high defect rate technologies. To this end, we empirically show that the classification accuracy of a baseline TPU drops significantly even at extremely low fault rates (as low as $0.006\%$). We then propose two novel strategies, fault-aware pruning (FAP) and fault-aware pruning+retraining (FAP+T), that enable the TPU to operate at fault rates of up to $50\%$, with negligible drop in classification accuracy (as low as $0.1\%$) and no run-time performance overhead. The FAP+T does introduce a one-time retraining penalty per TPU chip before it is deployed, but we propose optimizations that reduce this one-time penalty to under 12 minutes. The penalty is then amortized over the entire lifetime of the TPU's operation.

## 1. Introduction

Deep neural networks (DNN) have, in the past few years, surpassed the performance of traditional machine learning algorithms and are now considered as state-of-the-art for a range of applications, including image and video recognition, text classification [1], [2] and language translation [3]. One drawback of DNNs, however, is their complexity. State-of-the-art DNNs have millions of parameters and are computationally expensive, both to train and execute. GPU acceleration is one way to mitigate this computational burden. Nonetheless, there is a growing interest in *special purpose hardware accelerators* for DNN execution to achieve even greater performance and energy efficiency. Examples of recent special-purpose DNN accelerators include [4]–[7].

DNNs contain multiple layers of computation. Each layer comprises of a computationally expensive matrix multiplication or convolution operation, the latter for convolutional neural networks (CNN), followed by an inexpensive non-linear "activation." Work dating back to the 1980s has shown that matrix multiplications and convolutions can be efficiently implemented using *systolic arrays* [8]. A systolic array is a grid of connected processing elements (PE) that only communicate with their neighbors. For matrix multiplications and convolutions, each PE simply performs a multiply-and-accumulate (MAC) operation and passes the accumulated value downstream. Data is streamed through the array in a synchronized manner such that each PE received its inputs at just the right time, thus obviating the need for input buffering or complex routing. Further, the cost of reading inputs from memory is amortized over several compute cycles, providing high energy efficiency.

An example of a systolic array based DNN accelerator is the Google Tensor Processing Unit (TPU), that uses $256 \times 256$ grid of MAC units at its core, and provides between $30\times$ to $80\times$ times greater performance than CPU or GPU based servers [6]. The TPU is widely deployed in Google datacenters to accelerate DNN inference while reducing the total datacenter energy consumption. In this paper, we use the TPU architecture from Google as the baseline design, but our proposed techniques can apply to any systolic array based DNN accelerators. In the rest of the paper, we will use TPU to refer to the general class of systolic array based DNN accelerators.

An important challenge with future technology scaling is the increase in fault rates, including both permanent (hard errors) and temporary faults (soft errors) [9]–[12]. Temporary faults might occasionally impact the DNN's classification results, but their overall impact on classification accuracy is small, even at high soft error rates. On the other hand, permanent faults, as we show in this paper, can affect the result of *every* DNN execution and significantly reduce the classification accuracy. While permanent faults, at least those that are related to manufacturing defects, can be identified during post-fabrication testing, discarding every chip with a permanent fault reduces yield. This paper addresses the design of *fault-tolerant TPU*, with a focus on manufacturing defects or process variation induced permanent faults.

Prior work has suggested that several machine learning (ML) applications, including DNNs, are inherently resilient to errors; One potential solution for addressing permanent faults in a TPU, therefore, is to simply ignore the permanent faults in the hope that the application itself is inherently resilient to errors. Indeed, prior work has suggested that this is the case for several machine learning (ML) applications, including DNNs [13]. However, using detailed gate-level simulations of stuck-at permanent faults, we show that the accuracy of the TPU drops sharply even if as few as four out of the approximately 64K MAC units are faulty.

Next, we propose two new solutions: *fault-aware prun-*

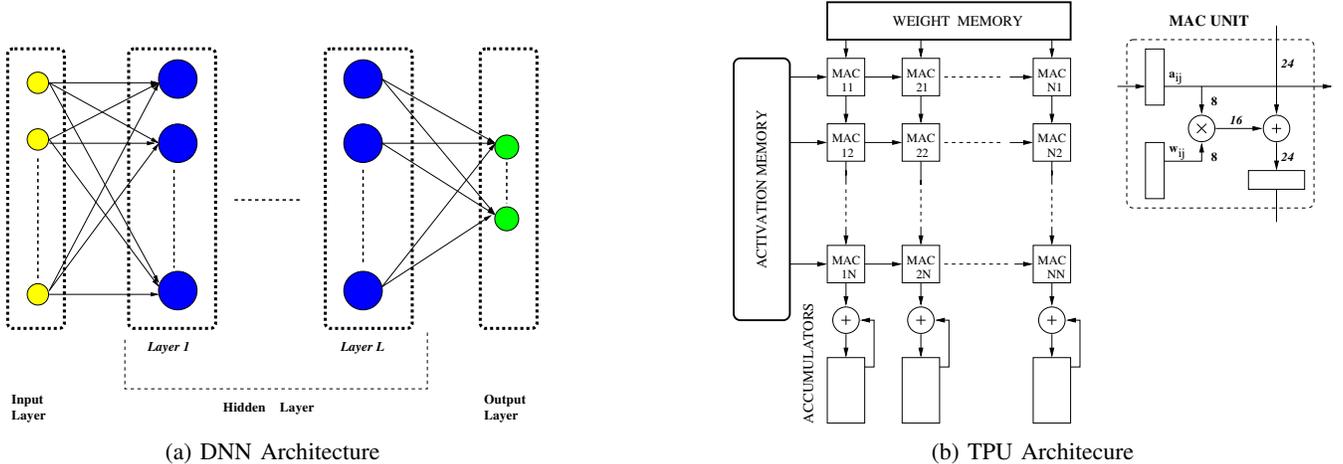

Figure 1: Illustration of DNN Architecture and TPU.

*ing* (FAP) and *fault-aware pruning plus retraining* (FAP+T) that enable the TPU to operate at high fault rates of up to 50% with a negligible impact on classification accuracy and no run-time performance overhead compared to the defect-free baseline. Our proposed solutions build on the recent work that shows that a significant fraction of a DNN's connections can be *pruned* with no (or limited) impact on accuracy. However, while the prior work using pruning to reduce DNN execution time and memory usage [14]–[18], we do so to enable *fault tolerance*. FAP prunes all connections in a DNN that map to faulty MACs using simple bypass circuitry that requires only minor modifications to the baseline TPU. FAP+T additionally *retrains* the DNN after pruning to restore classification accuracy back or close to its baseline, but comes at the expense of extra "test time" per TPU chip. To the best of our knowledge, ours is the *first* work that analyzes and proposes solutions to mitigate the impact of permanent faults on TPU (or TPU-like) DNN accelerators.

## 2. Related Work

Our work falls in the area of fault-tolerant digital system design. The prior work in this area has a rich history, starting with early work on addressing permanent faults in memory systems using error correcting codes (ECC) [19], and redundancy based approaches for addressing faults in logic [20]–[23] and in the routing fabric [24], [25]. However, these techniques do not directly apply to systolic arrays.

Kung et al. [26] were the first to describe fault-tolerant systolic array designs, which were later improved upon by [27]. The basic idea is to view a systolic array with faults as a smaller systolic array with *fewer* rows and columns. These solutions, in addition to requiring complex bypassing and additional registers inside PEs, can have a significant performance penalty at high-defect rates since an entire column/row is eliminated for each faulty PE. More sophisticated solutions can reduce the performance penalty, but at even higher design complexity [28], [29]. It is important to note that, as opposed to FAP and FAP+T (our proposed solutions), all of these techniques are *application agnostic*, that is, they seek to correctly execute the original task/application in the presence of faults. FAP and FAP+T, on the other hand, are application-aware, that is, we modify the underlying DNN architecture using pruning and re-training to adapt to faults in the TPU. Consequently, FAP and FAP+T have no performance penalty and negligible area overhead.

Another line of related work is on the design of crossbar-based DNN accelerators leveraging emerging technologies like phase change memory (PCM), resistive RAM (RRAM) and spin-based devices [30], [31]. Specifically, prior work has proposed techniques to cope with non-ideal device characteristics including device non-linearities, limited dynamic range and process variations. However, these techniques are specific to the devices used and to crossbar architectures, and do not apply to TPU designs [32], [33].

## 3. Background and Preliminaries

In this section, we breifly describe the relvant background on DNNs and the design of our baseline hardware accelerator for DNN execution.

### 3.1. Deep Neural Networks

A DNN consists of $L$ stacked layers of computation, as shown in Figure 1a. Layer $l$ has $N^l$ *neurons* whose outputs are referred to as *activations*, represented by an $N^l$ dimensional vector $a^l$. Each layer multiplies the vector of activations from the previous layer with a weight matrix $w^l$ of dimensions $N^{l-1} \times N^l$ and adds constant biases represented by an $N^l$ dimensional vector $b^l$, followed by an element-wise activation function $\phi$. Mathematically, each layer of a DNN performs the following operation:

$$y_i^l = \sum_j w_{i,j}^l a_j^{l-1} + b_i^l, \quad a_i^l = \phi(y_i^l) \quad \forall l \in [1, L], \quad (1)$$

In our description so far, we have assumed that each neuron's activation depends on the activations of all neurons in the previous layers. Such layers are referred to as *fully connected* (FC) layers. A commonly used special case of an FC layer is a convolutional (conv) layer. The activations of a

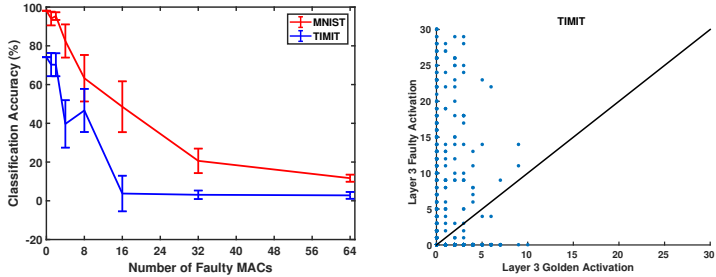

(a) Classification Accuracy Drop Due to Stuck-at-Fault MACs.

(b) TIMIT Output Regression for Layer 3 Activation with 8 Faulty MACs.

Figure 2: Impact of TPU Stuck-at-Faults on DNN Applications

convolutional neural network are represented as 3-D *tensors* of dimensions $W^l \times H^l \times D^l$ which refer to the width, height and number of channels in layer $l$ of the network. Also, the weights are represented as 4-D tensors of dimensions $F^l \times F^l \times D^{l-1} \times D^l$, where $F^l$ is the filter size in layer $l$. While the mathematical expression of the convolutional layer is omitted for brevity; for the purposes of this paper it suffices to know that the outputs of channel $d \in D^l$ is determined by convolving the 3-D input tensor with a 3-D weight tensor obtained by setting the last dimension of the 4-D weight tensor to $d$.

### 3.2. DNN Acceleration on TPU

Figure 1b shows a block-diagram of the TPU architecture, at the heart of which is a systolic array containing $N \times N$ MAC units that is used to perform the computationally expensive matrix multiplication and convolution operations. To understand how, consider a fully-connected layer with $N$ input neurons and $N$ output neurons, and consequently an $N \times N$ weight matrix. The weight matrix is first loaded into the systolic array, with weight $w_{i,j}$ being loaded into the MAC in row $j$ and column $i$, which we refer to as $MAC_{i,j}$.

Now consider the first column of the TPU. $MAC_{1,1}$ computes $w_{1,1}a_1$ in the first clock cycle and sends the result to $MAC_{1,2}$ which adds the product $w_{1,2}a_2$ to its input and forwards the result further downwards. In clock cycle $N$, the $MAC_{1,N}$ unit outputs $y_1 = \sum_{i=1}^{N} w_{1i}a_i$, which is the the first element of the output matrix.

The second column receives the same stream of inputs as the first column, but delayed by one clock cycle. This column is loaded with weights from the second row of the weight matrix, and outputs $y_2$, and so on. In practice, the TPU operates in *batches*, where each batch has $B$ inputs. A batch of $B$ inputs is multiplied by an $N \times N$ weight matrix in $2N + B$ clock cycles.

### 4. Motivational Analysis: Impact of Permanent Faults on TPU

To motivate the proposed fault-tolerant TPU design, we begin by empirically analyzing the impact of permanent faults on the classification accuracy of the TPU. To do so, we first synthesized an RTL description of the TPU into the gate-level netlist using the 45 nm OSU PDK, and then inserted stuck-at faults at internal nodes in the gate-level netlist. For this analysis, we focused only on faults in the data-path and ignored faults in the memory components (since they can be addressed using ECC) and the control logic since it consumes an insignificant fraction of the design.

We then mapped DNNs for two classification tasks, MNIST digit classification and TIMIT speech recognition, on the faulty TPU and plot the classification accuracy versus number of faulty MAC units in Figure 2a. For TIMIT, we observe that even with only four faulty MACs (i.e., with only $\sim 0.005\%$ MACs faulty), the classification accuracy drops from the $74.13\%$ to $39.69\%$.

The reason for the drop in accuracy is that stuck-at faults frequently affect the higher order bits of the MAC output, resulting in large absolute errors in the matrix-vector product. Figures 2b scatters the golden (fault-free) activations of the final layer of the TIMIT DNNs with the corresponding faulty outputs. Observe that for TIMIT, the faulty outputs have much higher magnitudes than the golden outputs.

To mitigate the permanent faults on the TPU, one simple idea is to bypass the entire columns where the faulty MACs reside. However, with the number of permanent faults increasing, the performance penalty would be unacceptable.

These observations together motivates the fault aware design of this paper.

### 5. Proposed Fault-Tolerant TPU Design

From the motivational example, it is clear that TPUs with even a relatively small number of faulty MAC units cannot be used unless the fault impact is mitigated. Our proposed fault-tolerant TPU design starts with an important observation: note from the description of the TPU operation in Section 3.2 that *each weight in the DNN maps to exactly one MAC unit*. In other words, there is a static mapping between DNN weights and MAC units. We then exploit the static mapping to determine which weights to prune.

This observation can be formalized using mapping functions $r()$ and $c()$ that take as input the indices of a DNN weight and output the row and column, respectively, of the MAC unit on which the weight is mapped. Specifically, based on the discussion in Section 3.2, the mapping functions for weight $w_{i,j}$ in a fully-connected layer are $r(i,j) = j\%N$ and $c(i,j) = i\%N$, where $\%$ is the modulo operator. Note that implicit in this mapping function is the fact that weight matrices that do not fit fully in the systolic array are first *blocked* into smaller $N \times N$ sub-matrices.

Our mapping strategy for conv layers sums along input channels along rows of the TPU, and each column computes outputs for a different output channel. Consequently, the mapping functions for weight $w_{i,j,k,l}$ in a conv layer are $r(i,j) = k\%N$ and $c(i,j) = l\%N$. We now describe our two proposed fault-tolerance methodologies: FAP and FAP+T.

## 5.1. Fault-Aware Pruning (FAP)

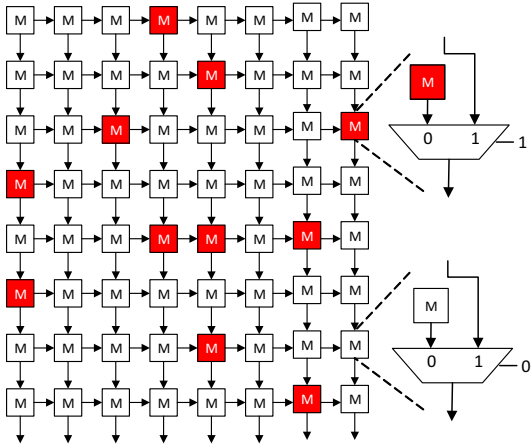

Figure 3: Diagram illustrating FAP.

The proposed FAP and FAP+T techniques both assume that standard post-fabrication tests are used on each TPU chip to determine the location of faulty MACs. Given this information, the idea behind FAP is to set (or prune) any weight that maps to a faulty MAC to zero. That is, for all pairs of $(i,j)$ values such that $\text{MAC}_{c(i,j),r(i,j)}$ is faulty, we set the corresponding $w_{i,j} = 0$ (this is for fully connected layers, a similar strategy is used for conv layers). Note that multiple weights can map to one MAC unit; correspondingly, even a single faulty MAC can result in multiple weights being pruned.

In hardware, pruning is achieved by introducing a separate bypass path for faulty MAC units, as shown in Figure 3. With the bypass path being enabled, the faulty MAC unit's contribution to the column sum is skipped, which is equivalent to setting the faulty MAC's weight to zero. Note that actually loading a zero weight into a faulty MAC is *not* equivalent to setting it's weight to zero. The area overhead due to the new bypass path is only about 9%.

## 5.2. FAP+T Solution

The FAP+T approach starts with FAP based on each TPU's fault map, but additionally *retrains* the unpruned weights in the DNN while forcing all pruned weights to zero during the re-training process. The FAP+T algorithm (see Algorithm 1) returns new, optimized values for the unpruned weights that improve the classification accuracy compared to the FAP solution. One drawback, however, is that re-training needs to be performed for each TPU chip based on its unique fault map; however, this needs to be done *only once* per TPU chip and the cost of doing so is amortized over the entire lifetime of the TPU chip.

Note that the FAP+T algorithm has a parameter $MAX\_EPOCHS$ that determines the number of iterations of the re-training algorithm. Setting this parameter to zero is equivalent to FAP. As $MAX\_EPOCHS$ is increased, the re-training time increases in return for increased classification accuracy.

**Algorithm 1: FAP+T Training Algorithm**

1 **Algorithm** FAP+T()
2 Load the pre-trained DNN weights and TPU fault map;
3 Determine indices of pruned weights from TPU fault map;
4 Set all pruned weights to zero;
5 **for** *Training epochs* $\leq MAX\_EPOCHS$ **do**
6 $\quad$ Update weights using back-prop.;
7 $\quad$ Set all pruned weights to zero;
8 **end**
9 **return** *Retrained model;*

| Multi-Layer Perceptrons (MLPs) | | |
|---|---|---|
| Name | MNIST | TIMIT |
| # Neurons | 784-256-256-256-10 | 1845-2000-2000-2000-183 |

| AlexNet Convolutional Feature Extraction | | | | |
|---|---|---|---|---|
| layer | filter | stride | padding | activation |
| conv1 | 96x3x11x11 | 4 | 0 | ReLU+LRN |
| pool1 | max, 3x3 | 2 | 0 | / |
| conv2 | 256x96x5x5 | 1 | 2 | ReLU+LRN |
| pool2 | max, 3x3 | 2 | 0 | / |
| conv3 | 384x256x3x3 | 1 | 1 | ReLU |
| conv4 | 384x384x3x3 | 1 | 1 | ReLU |
| conv5 | 256x384x3x3 | 1 | 1 | ReLU |
| pool5 | max, 3x3 | 2 | 0 | / |

| AlexNet Fully-connected Layers | | |
|---|---|---|
| layer | #neurons | activation |
| fc6 | 4096 | ReLU |
| fc7 | 4096 | ReLU |
| fc8 | 4096 | / |

TABLE 1: Benchmark DNNs architecture

## 6. Empirical Evaluation

In this section, we introduce our evaluation framework and discuss the results on three popular DNN applications.

### 6.1. Setup

**Benchmarks** The three benchmarks we used are: MNIST digit classification, TIMIT speech recognition [34] and AlexNet for image recognition using the PASCAL VOC2007 dataset [35]. We trained all DNNs from scratch and achieve the-state-of-art classification accuracy. Table 1 shows the detail parameters of the three DNN architectures. The MNIST and TIMIT DNNs have only fully connected layers, while AlexNet has five conv layers and three fully connected layers.

**Systolic Array Parameters** A prototype of the systolic array of a grid of $256 \times 256$ MACs was developed in synthesizable Verilog and synthesized with OSU FreePDK 45 $nm$ standard cell library using the Cadence Genus synthesis tool. The systolic array is synthesized to be run at 658 $MHz$ with a nominal supply voltage of $1.1V$ and consumes $19.7W$ of dynamic power.

**Simulation Methodology** We scheduled our benchmark DNNs on the systolic array using the mapping techniques described in Section 5. To analyze the impact of permanent faults on the TPU, we use stuck-at fault injection methodology described in Section 4. We then conduct a detailed gate-level simulation in Modelsim to collect the final accuracy for the benchmarks. Each experiment is repeated 10 times with faults injected in different locations, picked uniformly at random.

### 6.2. Results

Figure 4 shows the classification accuracy versus the number of permanent faults using the FAP and FAP+T techniques for our three benchmark DNNs. Note that we have about 65K MAC units in the TPU, and show results for high defect rates of up to 50%, i.e., with upto half of the MAC units faulty. For MNIST and TIMIT, both FAP and FAP+T are able to go up to fault rates of 25% with negligible classification accuracy reduction. However, with 50% faulty MACs, only FAP+T is able to provide close to the baseline classification accuracy, while that of the FAP technique reduces.

Compared to MNIST/TIMIT, the classification accuracy for AlexNet as seen in Figure 4b, drops markedly with increasing fault rates with FAP. FAP+T, however, performs much better and only incurs an 8% accuracy drop even with a 50% fault rate. One reason for the accuracy drop comes from the scheduling policy for the CNNs. In the current policy, one permanent faulty MAC would lead to a whole channel of the filter to be pruned. A more sophisticated scheduling may show better results but it is out of the scope of this paper.

As mentioned before, FAP+T incurs a one-time re-training overhead per TPU chip. While the retraining overhead (about 1 hour for AlexNet in the worst case, 25 epochs) is negligible compared to the lifetime of the TPU, we show in Figure 5 that it can be reduced even further by setting the value of $MAX\_EPOCHS$ appropriately. Specifically, Figure 5 shows the classification accuracy of FAP+T with increasing number of training epochs. Observe, for instance, that for AlexNet with 25% faulty MACs, the classification accuracy after the first 5 epochs only marginally lower than the classification accuracy after 25 epochs. Consequently, the re-training time for this example of AlexNet can be reduced by 5×, from 1 hour to only 12 minutes.

## 7. Conclusion

In this paper, we address the problem of designing fault-tolerant hardware accelerators for DNN execution for high defect rate technologies. Using the systolic array based Google TPU architecture as a baseline, we first showed using detailed gate-level simulations that the TPU's classification accuracy drops significantly even in the presence of very low fault rates. Then, we proposed two novel techniques, FAP and FAP+T, that allow TPUs to operate even with fault rates as high as 50%, with negligible to tolerable drops in classification accuracy (from 0.1% drop for TIMIT to 8% drop for AlexNet). While neither technique has any run-time performance overhead, FAP+T introduces a one-time retraining overhead of about 12 minutes per TPU chip for AlexNet. However, this one-time cost is amortized over the entire TPU's lifetime. As future work, we plan to address the impact of aging-related faults on DNN accelerators. Also, we are planning to apply our proposed methodology on larger neural-networks.

## References

[1] A. Krizhevsky *et al.*, "Imagenet classification with deep convolutional neural networks," in *Advances in Neural Information Processing Systems (NIPS)*, pp. 1097–1105, 2012.

[2] A. Karpathy *et al.*, "Large-scale video classification with convolutional neural networks," in *Proceedings of the IEEE conference on Computer Vision and Pattern Recognition*, pp. 1725–1732, 2014.

[3] I. Sutskever *et al.*, "Sequence to sequence learning with neural networks," in *Advances in Neural Information Processing Systems (NIPS)*, pp. 3104–3112, 2014.

[4] Y.-H. Chen *et al.*, "Eyeriss: An energy-efficient reconfigurable accelerator for deep convolutional neural networks," *IEEE Journal of Solid-State Circuits*, vol. 52, no. 1, pp. 127–138, 2017.

[5] Z. Du *et al.*, "Shidiannao: Shifting vision processing closer to the sensor," in *ACM SIGARCH Computer Architecture News*, vol. 43, pp. 92–104, 2015.

[6] N. P. Jouppi *et al.*, "In-datacenter performance analysis of a tensor processing unit," in *Proceedings of the 44th Annual International Symposium on Computer Architecture*, pp. 1–12, ACM, 2017.

[7] S. Park *et al.*, "4.6 a1. 93tops/w scalable deep learning/inference processor with tetra-parallel mimd architecture for big-data applications," in *IEEE International Solid-State Circuits Conference (ISSCC)*, pp. 1–3, 2015.

[8] H.-T. Kung, "Why systolic architectures?," *IEEE computer*, vol. 15, no. 1, pp. 37–46, 1982.

[9] S. S. Sahoo *et al.*, "Design and evaluation of reliability-oriented task re-mapping in mpsocs using time-series analysis of intermittent faults," in *Design, Automation & Test in Europe Conference & Exhibition (DATE)*, pp. 798–803, IEEE, 2016.

[10] J. Zhang *et al.*, "Enabling extreme energy efficiency via timing speculation for deep neural network accelerators," 2017.

[11] S. Borkar, "Design perspectives on 22nm cmos and beyond," in *Proceedings of the 46th Annual Design Automation Conference*, pp. 93–94, ACM, 2009.

[12] C. Constantinescu, "Trends and challenges in vlsi circuit reliability," *IEEE Micro*, vol. 23, no. 4, pp. 14–19, 2003.

[13] A. Gebregiorgis *et al.*, "Error propagation aware timing relaxation for approximate near threshold computing," in *Proceedings of the 54th Annual Design Automation Conference (DAC)*, pp. 1–6, IEEE, 2017.

[14] S. Han *et al.*, "Deep compression: Compressing deep neural networks with pruning, trained quantization and huffman coding," *arXiv preprint arXiv:1510.00149*, 2015.

[15] J. Yu *et al.*, "Scalpel: Customizing dnn pruning to the underlying hardware parallelism," in *Proceedings of the 44th Annual International Symposium on Computer Architecture*, pp. 548–560, ACM, 2017.

[16] H. Li *et al.*, "Pruning filters for efficient convnets," *arXiv preprint arXiv:1608.08710*, 2016.

[17] S. Anwar *et al.*, "Structured pruning of deep convolutional neural networks," *ACM Journal on Emerging Technologies in Computing Systems (JETC)*, vol. 13, no. 3, p. 32, 2017.

[18] P. Molchanov *et al.*, "Pruning convolutional neural networks for resource efficient inference," 2016.

[19] L. Levine *et al.*, "Special feature: Semiconductor memory reliability with error detecting and correcting codes," *Computer*, vol. 9, no. 10, pp. 43–50, 1976.

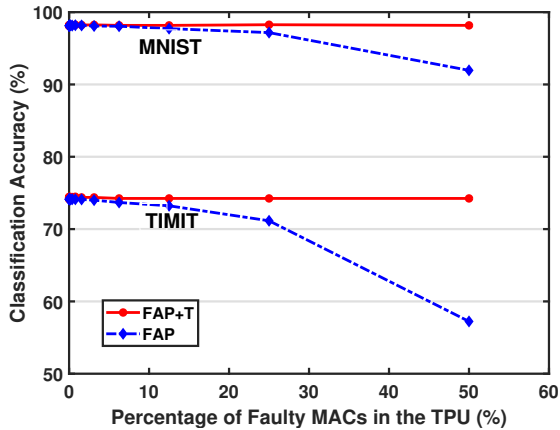
(a) MNIST and TIMIT.

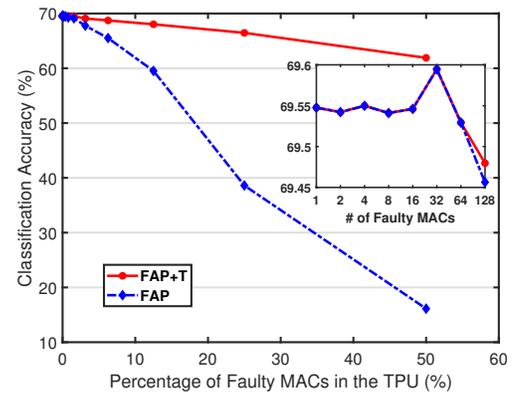
(b) AlexNet.

Figure 4: Classification accuracy vs. Percentage of Faulty MACs using FAP and FAP+T for (a) MNIST and TIMIT and (b) AlexNet.

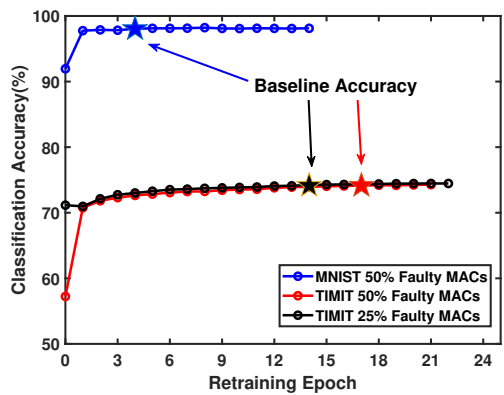
(a) MNIST and TIMIT.

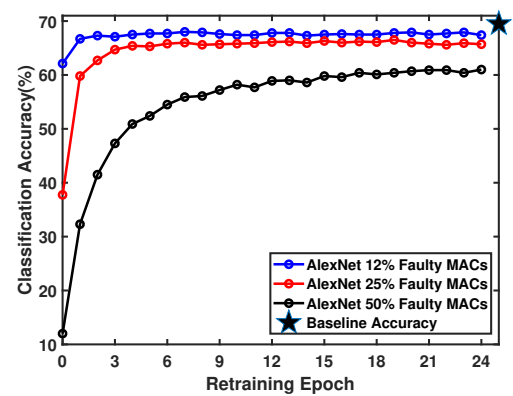
(b) AlexNet.

Figure 5: Classification accuracy vs. $MAX\_EPOCHS$ for (a) MNIST and TIMIT, and (b) AlexNet.

[20] J. H. Patel et al., "Concurrent error detection in alu's by recomputing with shifted operands," *IEEE Transactions on Computers*, vol. 31, no. 7, pp. 589–595, 1982.

[21] N. Oh et al., "Error detection by duplicated instructions in super-scalar processors," *IEEE Transactions on Reliability*, vol. 51, no. 1, pp. 63–75, 2002.

[22] A. Meixner et al., "Argus: Low-cost, comprehensive error detection in simple cores," in *40th Annual IEEE/ACM International Symposium on Microarchitecture (MICRO)*, pp. 210–222, 2007.

[23] M. Zhang et al., "Sequential element design with built-in soft error resilience," *IEEE Transactions on Very Large Scale Integration (VLSI) Systems*, vol. 14, no. 12, pp. 1368–1378, 2006.

[24] Y.-C. Chang et al., "On the design and analysis of fault tolerant noc architecture using spare routers," in *Proceedings of the 16th Asia and South Pacific Design Automation Conference (ASPDAC)*, pp. 431–436, 2011.

[25] W.-C. Tsai et al., "A fault-tolerant noc scheme using bidirectional channel," in *Proceedings of the 48th Design Automation Conference (DAC)*, pp. 918–923, ACM, 2011.

[26] H. Kung et al., "Fault-tolerance and two-level pipelining in vlsi systolic arrays," tech. rep., Carnegie-Mellon UNIV, 1983.

[27] J. H. Kim et al., "On the design of fault-tolerant two-dimensional systolic arrays for yield enhancement," *IEEE Transactions on Computers*, vol. 38, no. 4, pp. 515–525, 1989.

[28] M. Esonu et al., "Fault-tolerant design methodology for systolic array architectures," *IEE Proceedings-Computers and Digital Techniques*, vol. 141, no. 1, pp. 17–28, 1994.

[29] H. F. Li et al., "Restructuring for fault-tolerant systolic arrays," *IEEE Transactions on Computers*, vol. 38, no. 2, pp. 307–311, 1989.

[30] G. W. Burr et al., "Neuromorphic computing using non-volatile memory," *Advances in Physics: X*, vol. 2, no. 1, pp. 89–124, 2017.

[31] C. Liu et al., "Rescuing memristor-based neuromorphic design with high defects," in *Proceedings of the 54th Annual Design Automation Conference (DAC)*, pp. 87:1–87:6, ACM, 2017.

[32] G. W. Burr et al., "Experimental demonstration and tolerancing of a large-scale neural network (165 000 synapses) using phase-change memory as the synaptic weight element," *IEEE Transactions on Electron Devices*, vol. 62, no. 11, pp. 3498–3507, 2015.

[33] D. Chabi et al., "Robust neural logic block (nlb) based on memristor crossbar array," in *Proceedings of the IEEE/ACM International Symposium on Nanoscale Architectures*, pp. 137–143, 2011.

[34] J. Ba et al., "Do deep nets really need to be deep?," in *Advances in Neural Information Processing Systems (NIPS)*, pp. 2654–2662, 2014.

[35] M. Everingham et al., "The PASCAL Visual Object Classes Challenge 2007 (VOC2007) Results." http://www.pascal-network.org/challenges/VOC/voc2007/workshop/index.html.